\documentclass[sigconf]{acmart}
\renewcommand\footnotetextcopyrightpermission[1]{} 
\usepackage{cleveref}
\usepackage{enumitem}
\crefname{figure}{Fig}{Figs}
\crefname{table}{Table}{Tables}
\AtBeginDocument{%
  }

\begin{document}

\title{PostureObjectstitch: Anomaly Image Generation Considering Assembly Relationships in Industrial Scenarios}


\author{Zebei Tong}
\authornotemark[1]
\affiliation{%
  \institution{Beijing Institute of Technology}
  \city{Beijing}
  \country{China}
}

\author{Hongchang Chen}
\authornotemark[1]
\affiliation{%
  \institution{The Hong Kong Polytechnic University}
  \city{Hong Kong}
  \country{China}
}

\author{Yujie Lei}
\affiliation{%
  \institution{Beijing Institute of Technology}
  \city{Beijing}
  \country{China}
}

\author{Gang Chen}
\affiliation{%
  \institution{Beijing Institute of Technology}
  \city{Beijing}
  \country{China}
}

\author{Yushi Liu}
\affiliation{%
  \institution{Beijing Institute of Technology}
  \city{Beijing}
  \country{China}
}

\author{Zhi Zheng}
\affiliation{%
  \institution{Li Auto}
  \city{Beijing}
  \country{China}
}

\author{Hao Chen}
\affiliation{%
  \institution{Li Auto}
  \city{Beijing}
  \country{China}
}

\author{Jieming Zhang}
\affiliation{%
  \institution{Li Auto}
  \city{Beijing}
  \country{China}
}

\author{Ying Li}
\affiliation{%
  \institution{Beijing Institute of Technology}
  \city{Beijing}
  \country{China}
}

\author{Dongpu Cao}
\affiliation{%
  \institution{Tsinghua University}
  \city{Beijing}
  \country{China}
}

\renewcommand{\shortauthors}{Trovato et al.}

\begin{abstract}
Image generation technology can synthesize condition-specific images to supplement real-world industrial anomaly data and enhance anomaly detection model performance. Existing generation techniques rarely account for the pose and orientation of industrial components in assembly, making the generated images difficult to utilize for downstream application. To solve this, we propose a novel image synthesis approach, called PostureObjectStitch, that achieves accurate generation to meet the requirement of industrial assembly. A condition decoupling approach is introduced to separate input multi-view images into high-frequency, texture, and RGB features. The feature temporal modulation mechanism adapts these features across diffusion model time-steps, enabling progressive generation from coarse to fine details while maintaining consistency. To ensure semantic accuracy, we introduce a conditional loss that enhances critical industrial elements and a geometric prior that guides component positioning for correct assembly relationships. Comprehensive experimental results on the MureCom dataset, our newly contributed DreamAssembly dataset, and the downstream application validate the outstanding performance of our method.
\end{abstract}




\begin{teaserfigure}
  \includegraphics[width=\textwidth]{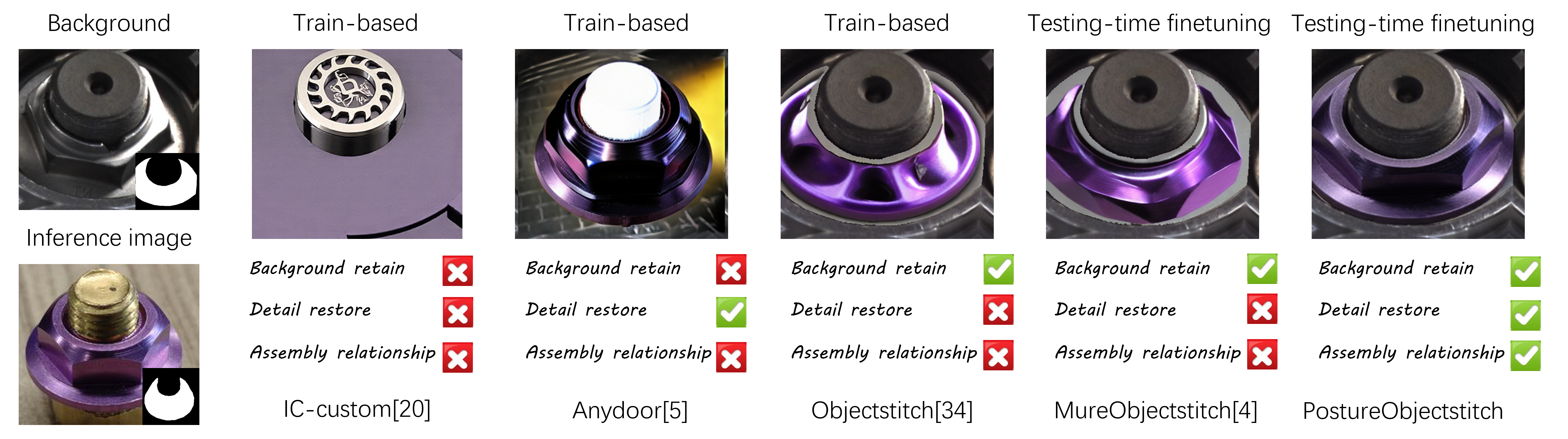}
  \caption{Fantastic application of our proposed PostureObjectstitch in industrial anomaly generation considering assembly relationships. Compared to other methods, we simultaneously consider background retention, detail restoration, and assembly relationships.}
  \Description{Enjoying the baseball game from the third-base
  seats. Ichiro Suzuki preparing to bat.}
  \label{fig:teaser}
\end{teaserfigure}

\received{20 February 2007}
\received[revised]{12 March 2009}
\received[accepted]{5 June 2009}

\maketitle

\section{Introduction}
Anomaly image generation is a critical approach to addressing challenges in industrial scenarios, such as data scarcity and sample imbalance, and plays a vital role in enhancing the performance of downstream anomaly detection models \cite{st01,st02,st03,st04,st05,st06,st07}. 
In industrial production, anomalous samples such as component misassembly are extremely rare, yet they can cause significant economic losses and safety hazards when they occur.
Traditional data collection methods are not only costly but also difficult to cover all possible anomalous scenarios. 
With the rapid development of generative models, synthesizing high-quality anomalous images through image composition techniques provides a novel solution to address this problem.

In general domains, image synthesis methods represented by diffusion models have achieved remarkable progress.
From text-to-image works like SDXL \cite{podell2023sdxlimprovinglatentdiffusion}, ControlNet \cite{controlnet}, and DALL-E 3 \cite{dall-e3}, to image-to-image methods such as InstructPix2Pix \cite{brooks2023instructpix2pixlearningfollowimage} and CustomDiffusion \cite{customdifussion}, and object composition approaches including AnyDoor \cite{anydoorzeroshotobjectlevelimage}, ObjectStitch \cite{objectstitchgenerativeobjectcompositing}, and Paint-by-Example \cite{paintexampleexemplarbasedimage}, these methods demonstrate powerful capabilities in natural scene image generation. 
However, these general approaches face fundamental adaptation challenges in industrial scenarios.

Industrial anomaly detection requires generated components to maintain their original geometric features and visual details. 
This is crucial for ensuring the accuracy of detection algorithms.
Existing text-guided generation methods can understand semantic descriptions. 
However, they cannot accurately restore fine visual features of specific components, such as thread textures and dimension markings.
Additionally, current generation methods lack effective adjustment mechanisms for target poses, making it difficult to modify component spatial positions and poses according to specific assembly scenario requirements.
These limitations prevent general image synthesis techniques from being directly applied to industrial anomaly image generation tasks.

To fill this gap, we propose the PostureObjectstitch which is a anomaly image generation mathod considering assembly
relationships in industrial scenarios. First, we design a condition decoupling strategy that decomposes input conditions into high-frequency (contour-focused), texture (detail-emphasizing), and RGB (global-information-capturing) features to better adapt to industrial generation requirements. Second, we propose a feature temporal modulation mechanism that enables decoupled features to function appropriately across different timesteps of diffusion models, ensuring progressive generation from coarse contours to fine textures.
Finally, we introduce a geometric prior and conditional loss, where the former guides accurate component positioning in target scenes to ensure correct assembly relationships, and the latter enhances generation quality of critical industrial information such as text.
Additionally, we construct the DreamAssembly dataset specifically for evaluating anomaly image generation in industrial assembly scenarios.
Experimental results demonstrate that our method generates high-quality industrial anomaly images and effectively addresses the limitations of general methods in assembly relationship modeling and detail fidelity.

The key technical contributions of our work are:
\begin{itemize}[noitemsep,topsep=0pt]
\item We propose a feature decoupling strategy that splits input conditions into high-frequency, texture, and RGB features, coupled with a time modulation mechanism for adaptive feature distribution across diffusion time-steps, enabling progressive generation from coarse to fine details.
\item We introduce a geometric prior to guide precise component positioning for correct assembly relationships and the conditional loss to ensure semantic consistency and assembly accuracy. 
\item We build the DreamAssembly dataset for industrial assembly anomaly generation evaluation and conduct experiments on both MureCom and our dataset. Results show our method maintains general scene compatibility while producing high-quality industrial anomaly images with assembly relationships.
\end{itemize}

\label{sec:intro}

\section{Related Work}
\subsection{Training-based diffusion models}

Pre-trained diffusion models, trained on large-scale image-text datasets, offer robust foundational capabilities for image generation by learning unified representations across diverse generation tasks.
GLIDE \cite{nichol2022glidephotorealisticimagegeneration} pioneers the application of diffusion models to text-guided image synthesis, establishing the foundational guidance mechanisms. Building on this, DALL-E2 \cite{ramesh2022hierarchicaltextconditionalimagegeneration} introduces the unCLIP framework, achieving remarkable semantic alignment. Imagen \cite{saharia2022photorealistictexttoimagediffusionmodels}  emphasizes the importance of large language models, leveraging T5 encoders to achieve exceptional generation quality.
A significant milestone is Stable Diffusion \cite{rombach2022highresolutionimagesynthesislatent}, which performs diffusion in latent space, significantly reducing computational requirements. Subsequent advancements by SDXL \cite{podell2023sdxlimprovinglatentdiffusion} and DALL-E 3 \cite{dall-e3} further improve image resolution and text adherence. In the realm of image editing, Paint-by-Example \cite{paintexampleexemplarbasedimage} enables reference-guided editing by conditioning diffusion models on CLIP-encoded exemplar images through cross-attention mechanisms. ObjectStitch \cite{objectstitchgenerativeobjectcompositing} achieves zero-shot object composition through seamless object transfer with automatic adaptation to target environments' lighting and perspective. Similarly, AnyDoor \cite{anydoorzeroshotobjectlevelimage} supports inserting arbitrary objects at any location while preserving their identity features.
However, these methods are inadequate for generating customized industrial compositions that demand precise pose and orientation control, as they frequently incur information loss in either foreground objects or background contexts.
\begin{figure*}[h]
    \centering
    \includegraphics[width=1\linewidth]{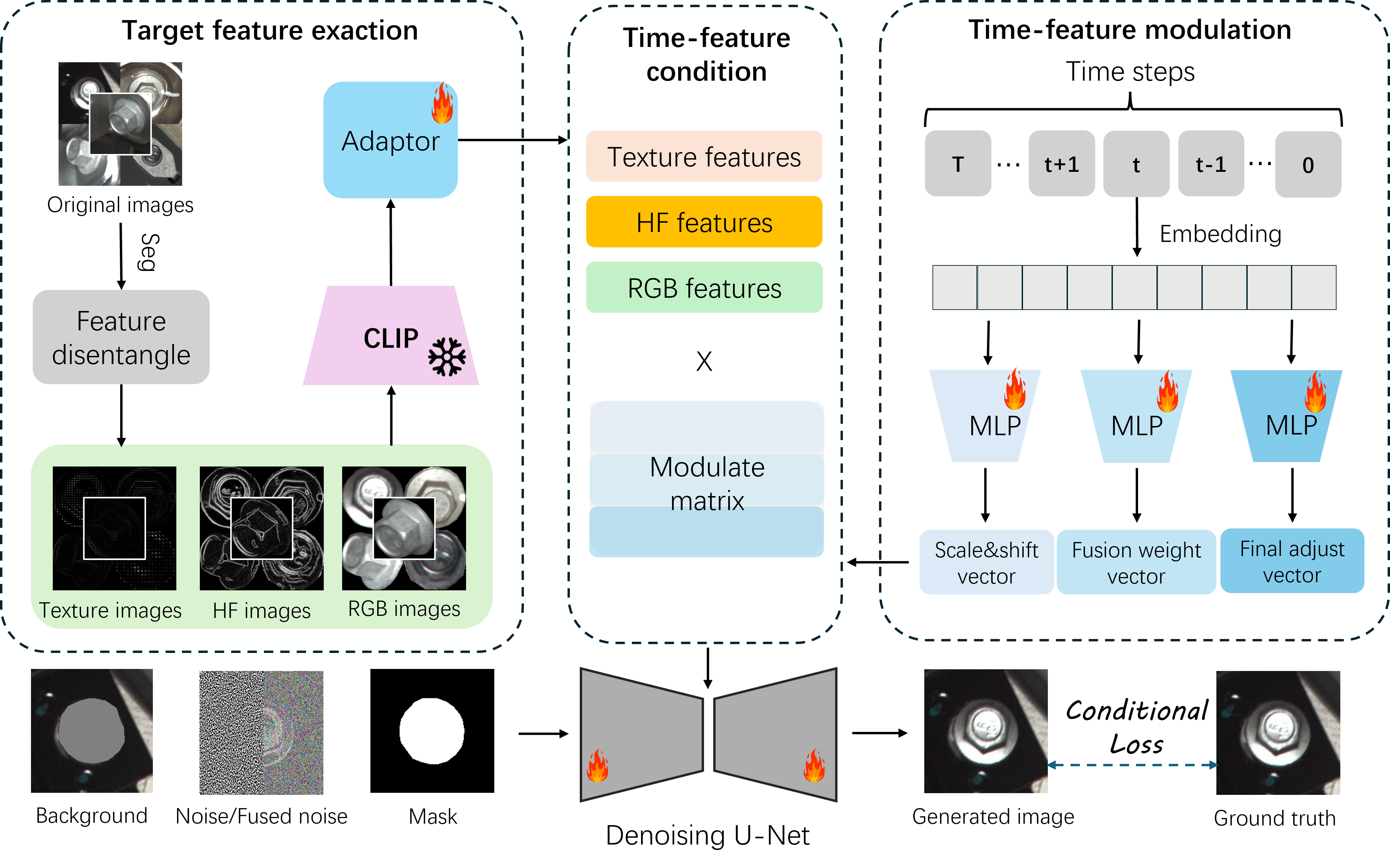}
    \caption{Overview of PostureObjectstitch. 
    Given N reference images of a specific sample, PostureObjectstitch fine-tunes the model to generate high-quality anomaly images for industrial assembly scenarios. Our method has three key components: First, we design a condition decoupling strategy that splits input information into high frequency features, texture features, and RGB features to better suit industrial generation needs. Second, we propose a feature temporal modulation mechanism that allows different features to work properly across diffusion timesteps, enabling gradual generation from coarse to fine details. Finally, we introduce positional pose priors and conditional loss. The former ensures accurate component positioning and correct assembly relationships, while the latter improves generation quality of industrial information like text.}
    \label{fig:placeholder}
\end{figure*}
\subsection{Testing-Time Finetuning}
Testing-time finetuning methods constitute a fundamental paradigm for personalized image generation, where pre-trained model parameters are adaptively optimized for specific target subjects during inference to achieve high-fidelity customized image synthesis.
Textual Inversion \cite{gal2022imageworthwordpersonalizing} first introduced the concept of optimizing the embeddings of learnable tokens by incorporating new tokens to represent target objects.
DVAR \cite{voronov2023lossinformativefastertexttoimage} improves training efficiency by proposing clear convergence criteria through removing randomness;
P+ \cite{pextendedtextualconditioning} and NeTI \cite{neuralspacetimerepresentationtexttoimage} enhance representation capability by employing distinct learnable tokens across different U-Net layers and introducing neural mappers for adaptive token embedding output respectively; 
HiFiTuner \cite{hifitunerhighfidelitysubjectdriven} integrates multiple techniques including mask-guided loss and parameter regularization; 

Although these methods have achieved promising results, token-based approaches inherently suffer from the fundamental limitation of detail loss, as they inevitably compress rich visual semantics into a limited set of learnable parameters.

Another category of fine-tuning methods focuses on adapting model parameters themselves to address the expressiveness limitations of token-based approaches.
DreamBooth \cite{ruiz2023dreamboothfinetuningtexttoimage} fine-tunes the entire diffusion model using rare token identifiers and class-specific regularization datasets to achieve high-quality subject personalization. 

To address storage efficiency, Custom Diffusion \cite{customdifussion} selectively fine-tunes key-value projections in cross-attention layers, balancing visual fidelity with reduced memory requirements.
Perfusion \cite{tewel2024keylockedrankeditingtexttoimage} extends cross-attention fine-tuning by regularizing key projections toward super-category embeddings and value projections toward learnable tokens. 
Mureobjectstitch \cite{mureobjectstitchmultireferenceimagecomposition} fine-tunes the model using multiple images of a specific subject, enabling multi-view customized generation.
Parameter-efficient methods including adapters \cite{xiang2023closerlookparameterefficienttuning, sohn2023styledroptexttoimagegenerationstyle, ye2023ipadaptertextcompatibleimage} and LoRA variants \cite{gu2023mixofshowdecentralizedlowrankadaptation,ruiz2024hyperdreamboothhypernetworksfastpersonalization,kong2024omgocclusionfriendlypersonalizedmulticoncept,lu2024objectdrivenoneshotfinetuningtexttoimage}have gained popularity for lightweight personalization.

Overall, parameter fine-tuning methods better preserve visual characteristics of customized subjects but suffer from pose accuracy limitations.
While mask guidance provides partial improvement, it fails when significant pose variations exist between generated and reference configurations.

\section{Method}

Given a collection of 3-5 multi-perspective images depicting a target foreground object, our approach leverages these reference views to generate anomaly images while explicitly modeling assembly-level spatial relationships. The core pipeline consists of three key components: target feature distanglement, time-feature modulation, and semantic auxiliary preservation. The original multi-view images are segmented first and disentangled into multiple features, which are then fed into the CLIP \cite{clip} encoder for feature extraction. Secondly, distinct feature types are adaptively modified according to time-steps, ensuring optimal utilization of decoupled features at each stage. Finally, to guarantee semantic consistency in the generation results, we introduce two complementary mechanisms: a conditional loss function that enforces textural coherence, and a position-pose prior mechanism that maintains geometric plausibility. Figure 3 depicts the comprehensive training and inference pipeline of the proposed framework.

\subsection{Target feature distanglement}
Following established image synthesis methodologies \cite{r00,r01,r02,iccustom,r03,dong2025dreamartistcontrollableoneshottexttoimage}, we employ a large-scale pre-trained image encoder as our target feature extraction backbone. We select CLIP as our encoder due to its demonstrated effectiveness in cross-modal representation learning. To best preserve the distinctive characteristics of target objects, we implement the following feature preservation operations:

\noindent \textbf{Foreground segmentation.} To minimize background interference with foreground object features, we first segment the foreground objects from the image background. The segmentation results are then placed within their minimum bounding rectangles, with the remaining areas filled with black pixels. This preprocessing step effectively removes most background information from the foreground images.

\noindent \textbf{Feature disentanglement.} Inspired by \cite{zhao2025hunyuan3d}, we disentangle the foreground features into high-frequency features and texture features, which are then fed into CLIP for feature extraction and encoding along with the original features obtained from the previous segmentation step. 

For high-frequency features, the processing pipeline is formulated as 
\cref{eq:1}.
\begin{equation}
F_{HF} = R(N( \alpha\cdot S(I) +  \beta\cdot L(I) + \gamma \cdot C(I))),
\label{eq:1}
\end{equation}
where $I$ represents the input image (automatically handles RGB to grayscale conversion), $S(\cdot)$ is the Sobel operator, $L(\cdot)$ is the Laplacian operator, $C(\cdot)$ means the Canny operator, $N(\cdot)$ is the normalization enhancement, $R(\cdot)$ shows output format conversion, $\alpha$, $\beta$, and $\gamma$ are hyperparameters.

For texture features, the processing pipeline is formulated as \cref{eq:2}.
\begin{equation}
F_{texture} = R(E(H(I))),
\label{eq:2}
\end{equation}
where $H(\cdot)$ represents the HOG \cite{hog} feature extraction and visualization operator, $E(\cdot)$ is the intensity rescaling enhancement operator, and $R(\cdot)$ is the RGB three-channel replication operator.
\begin{figure}[hp!]
    \centering
    \includegraphics[width=0.95\linewidth]{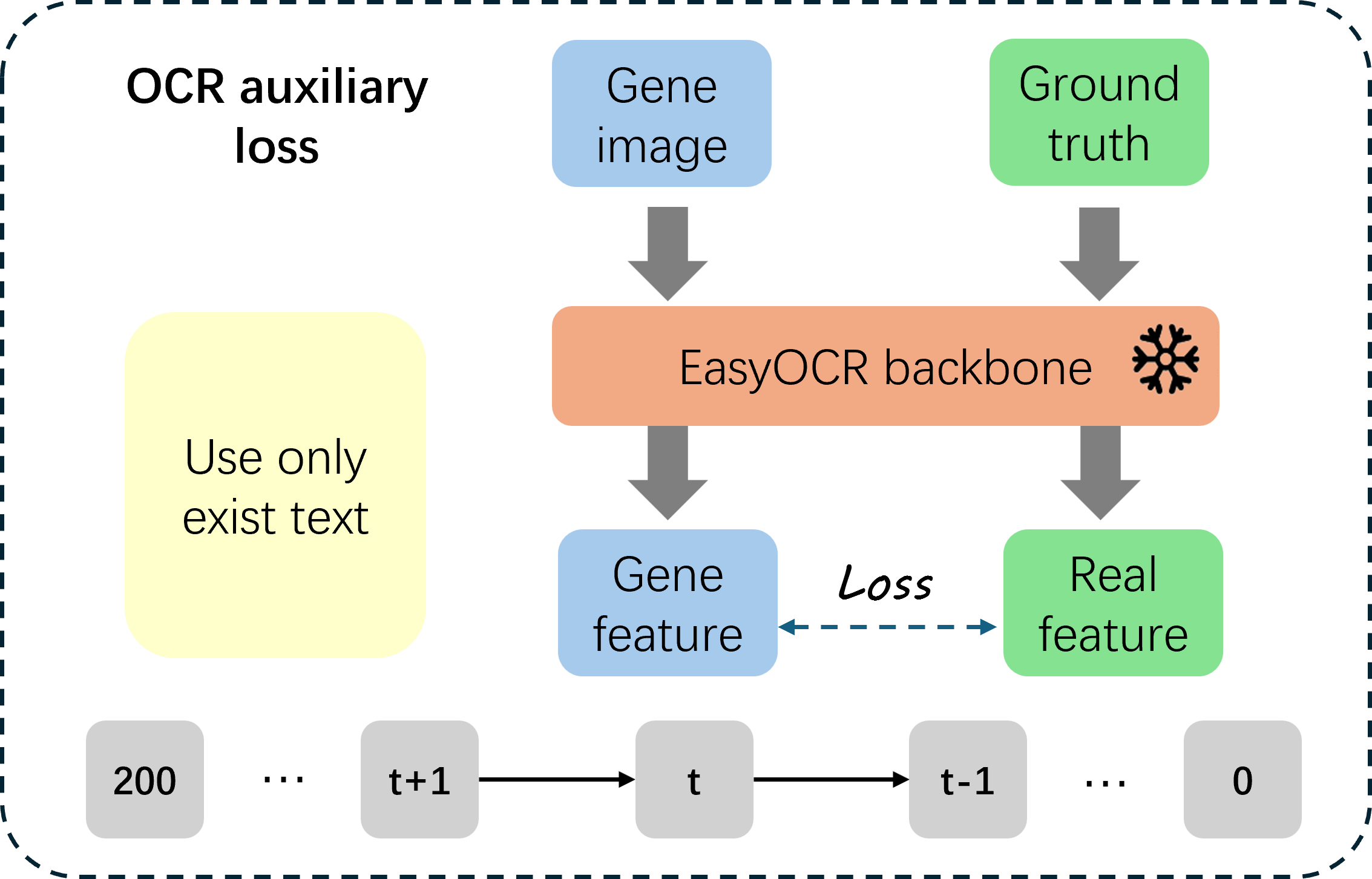}
    \caption{OCR auxiliary loss.}
    \label{ocrloss}
\end{figure}

\subsection{Time-feature modulation}
Based on the consensus from prior work \cite{rr00,rr01,rr02,rr03,xiao2023comcatefficientcompressioncustomization}, the image generation process of diffusion models follows a coarse-to-fine paradigm. As demonstrated in \cite{decatur2025reusing}, features at different hierarchical levels play distinct roles across different timesteps. Motivated by this observation, we introduce a time-feature modulation module that associates timestep information with the decoupled features, enabling adaptive scheduling of different features according to the current timestep.

The detailed pipeline is as follows. First, the timestep information is encoded into a temporal feature vector. Subsequently, this temporal feature vector is fed into both the feature modulation parameter generation module and the fusion weight generation module to obtain feature adjustment parameters.Finally, the timestep information is embedded into the features through a final modulation module. The feature adjustment process is formulated as shown in \cref{eq:3,eq:4,eq:5,eq:6,eq:7,eq:8}.
{\setlength{\jot}{10pt}
\begin{gather}
\mathbf{h}_t = \text{MLP}(\text{SinCos}(t, 320)),
\label{eq:3} \\
{\tiny [\mathbf{s}_{k}, \mathbf{b}_{k},] = \text{SSVM}(\mathbf{h}_t) , \quad k \in \{RGB, HF, Texture\}},
\label{eq:4} \\
\mathbf{F}_k^{\text{mod}} = \mathbf{F}_k \odot (\alpha \cdot \mathbf{s}_k) + \beta \cdot \mathbf{b}_k,
\label{eq:5} \\
[w_{k}] = \text{Softmax}(\text{FWVM}(\mathbf{h}_t)),
\label{eq:6} \\
\mathbf{F}^{\text{fused}} = w_{k} \cdot \mathbf{F}_{k}^{\text{mod}}, 
\label{eq:7} \\
\mathbf{F}^{\text{out}} = \mathbf{F}^{\text{fused}} +  \text{FAVM}(\mathbf{h}_t),
\label{eq:8}
\end{gather}
}
where $t$ is the timestep vector, \text{SinCos} means sinusoidal positional encoding, SSVM , FWVM and FAVM are scale and shift vector multilayer perceptron, fusion weight vector multilayer perceptron and final adjust vector multilayer perceptron,
 $\mathbf{s}_k$ and $\mathbf{b}_k$ are scale parameters and shift parameters for each feature type $k$.
 $\alpha$ and $\beta$ are hyperparameters.
$\mathbf{F}_k$ are the original features, $\odot$ denotes element-wise multiplication
$\mathbf{w}_k$ is fusion weight for each feature type $k$
and $\mathbf{F}_k^{\text{mod}}$ is the modulated features.
$\mathbf{F}^{\text{fused}}$ is the time-aware fused feature representation.
$\mathbf{F}^{\text{out}}$ represents the output condition.

\subsection{Semantic auxiliary preservation}

\textbf{Conditional loss.} In experiments involving foreground object insertion and replacement, we observe that current methods achieve satisfactory generation results for foreground contours and colors, but exhibit poor performance in generating textual information within foreground objects.
The semantic information contained in text is completely lost during the generation process.

To address this issue, we introduce a conditional loss function to constrain the consistency of text information generation.
As shown in \cref{eq:9}, 
\begin{equation}
L_{cond} = L_{VLB} + L_{MSE} + \mathbb{I}_{\text{text}} \cdot L_{OCR},
\label{eq:9}
\end{equation}
where $L_{\text{VLB}}$ is the variational lower bound loss that serves as the core loss of diffusion models for training the denoising process, $L_{\text{MSE}}$ is the mean squared error loss measuring pixel-level reconstruction error between generated and target images, $\mathbb{I}_{\text{text}}$ is a binary indicator function that equals 1 when the foreground object contains textual information and 0 otherwise, $L_{\text{OCR}}$ is the OCR auxiliary loss that constrains semantic feature consistency between generated and ground truth text.


The details of OCR auxiliary loss are presents in \cref{ocrloss}. 
Text content generation primarily occurs during the low-noise denoising phase. 
Based on this characteristic, we introduce OCR auxiliary loss during the denoising process when timestep t \textless 200.
During computation, we first reconstruct the denoised image from the diffusion model's prediction, then decode the reconstructed image from latent space to image space.
Subsequently, both the generated reconstructed image and the ground truth reference image are fed into a frozen OCR feature extraction backbone.
Finally, we compute a MSE loss on the extracted features.
\begin{figure}[hp!]
    \centering
    \includegraphics[width=1\linewidth]{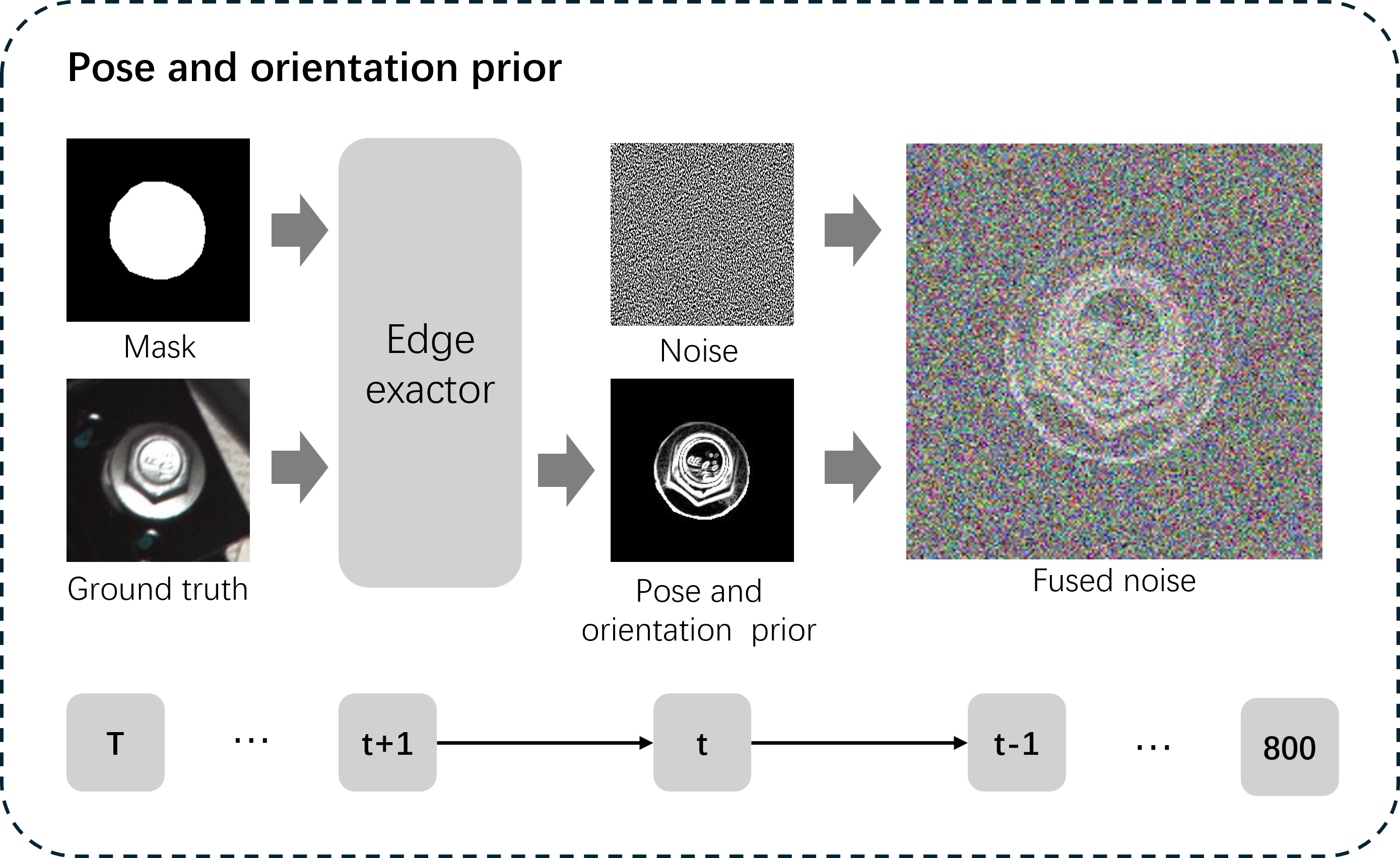}
    \caption{Pose and orientation prior fusion.}
    \label{pose}
\end{figure}
\noindent\textbf{Pose and orientation prior information.} In image replacement tasks, most existing methods focus solely on the consistency of object morphology and category before and after replacement, while neglecting the consistency of geometric information, which is crucial for maintaining assembly relationships in industrial scenarios.
To address this issue, we introduce pose and orientation prior information during both training and inference. 

The detailed implementation of Pose and orientation prior information is shown in \cref{pose} and the underlying principle is as follows: during training phases with high noise ratios, we extract high-frequency features from the foreground regions to be generated and superimpose these high-frequency features with noise. This operation provides positional and pose prior information during the denoising process from noise to image. For replacement task inference, we introduce the original pose and orientation information by superimposing the high-frequency features of the foreground to be replaced with the initial noise, thereby preserving the original pose and orientation information and ensuring consistency of assembly relationships.

\begin{figure*}
    \centering
    \includegraphics[width=1\linewidth]{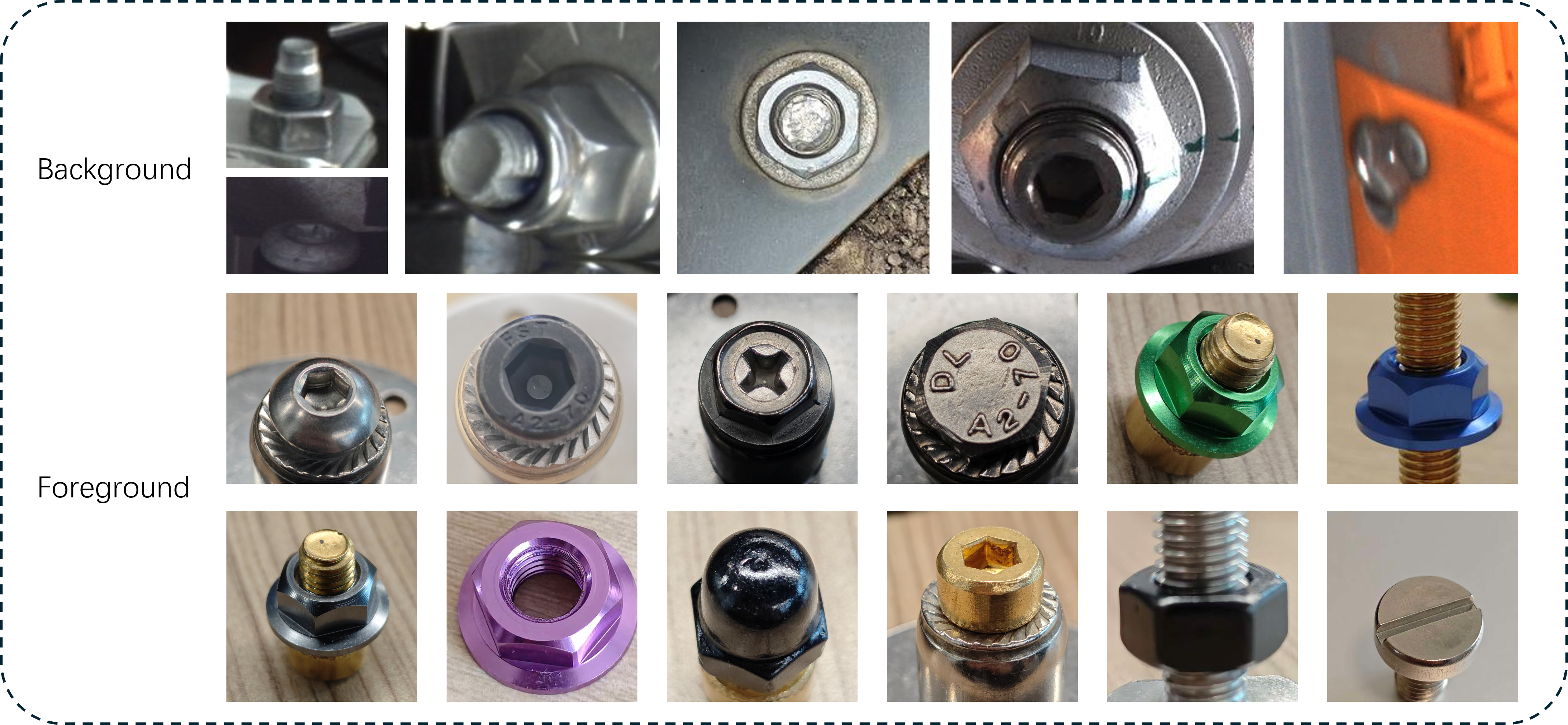}
    \caption{Dreamassembly dataset overview. \textbf{Background:}The background images are collected from real industrial environments on actual production lines. These backgrounds cover typical automotive assembly scenarios including assembly workstations, inspection platforms, and industrial equipment. \textbf{Foreground:} The foreground section presents 25 common key components in industrial assembly. Each component is captured under various poses including different rotation and tilt angles.}
    \label{datasetoverview}
\end{figure*}
\section{Dataset}
\textbf{MureCom} \cite{murecom}. This is a large-scale dataset specifically designed for image composition tasks. It encompasses 32 object categories, with each category providing 20 background images and 3 foreground object instances. The distinctive feature of this dataset lies in providing 5 reference images from different viewpoints for each foreground object, along with precise target placement region annotations for all background images. 

\noindent\textbf{DreamAssembly.} While numerous datasets exist for industrial anomaly detection, they provide limited focus on assembly relationship anomalies. To properly characterize and evaluate such problems, we collect and propose DreamAssembly, a dataset specifically designed for industrial assembly relationships.The overview of this dataset is shown in \cref{datasetoverview}. We partition assembly scenes into background and foreground components, conducting data collection for each part separately.
For backgrounds, we collect 500 authentic assembly backgrounds from over 50 inspection points on actual production lines, encompassing diverse viewpoints, lighting conditions, and resolutions to maximally replicate real manufacturing environments.
For foregrounds, we select 25 common assembly components and capture them under various poses and assembly states, providing fine-grained mask annotations for all foreground objects.

\section{Implementation Details}
\subsection{Experimental Setup}
Our method is built upon the pre-trained Stable Diffusion v1.4 model as the base framework.
To preserve the robust generative priors and reduce computational overhead, we freeze the parameters of the VAE (Variational Autoencoder) and the CLIP encoder during the training process, focusing our optimization entirely on the U-Net and the newly proposed modules.
All experiments are conducted with an input image resolution of $512 \times 512 pixels$. For model training, we employ the AdamW optimizer with a learning rate of $1 \times 10^{-5}$and a batch size of 8. The entire training process is performed on a single NVIDIA A100 GPU.

\subsection{Evaluation Metrics}
To rigorously assess the fidelity of the generated foreground objects, we adopt the established evaluation protocols from DreamBooth and DreamEdit \cite{dreamedit}.
Specifically, we employ two widely recognized metrics: DINO and CLIP-I.
DINO evaluates the structural and fine-grained feature consistency, while CLIP-I measures the high-level semantic alignment between the generated objects and the reference images.
To account for the inherent stochasticity of diffusion models and ensure statistical reliability, our evaluation process involves generating four distinct synthesized images for each specific foreground-background combination.
For every generated sample, we isolate the foreground elements and compute similarity scores by comparing them against the corresponding foreground components extracted from all available reference images. 
These individual similarity scores are subsequently averaged and aggregated, yielding a robust and comprehensive fidelity measure for each unique foreground-background pairing.

To complement the foreground analysis and ensure the integrity of the original scene, we quantitatively measure the background preservation quality. 
This is achieved through the computation of the Structural Similarity Index Measure (SSIM) \cite{ssim} and the Learned Perceptual Image Patch Similarity (LPIPS) \cite{lpips} metric. 
SSIM evaluates the low-level structural and luminance consistency, whereas LPIPS assesses the deep perceptual similarity. 
Crucially, to prevent the newly composed foreground objects from skewing the results, these computations are performed exclusively on the masked background regions of the synthesized outputs, directly comparing them to the original background images. 
This strict isolation guarantees that the metrics accurately reflect the model's ability to preserve the original background context without unintended alterations.


\subsection{Comparisons with baselines}
To comprehensively evaluate the effectiveness of our proposed approach, we compare our method against several representative baselines, specifically ObjectStitch, MureObjectStitch, AnyDoor, and IC-Custom.
During the evaluation process, we adopt distinct testing strategies based on the training paradigms of these methods.
For pre-trained approaches, including ObjectStitch, AnyDoor, and IC-Custom, we directly employ their officially provided pre-trained weights for evaluation without any additional training.
Conversely, for fine-tuning-based methods like our PostureObjectStitch and MureObjectStitch, we utilize the specifically fine-tuned weights during the assessment to accurately reflect their performance.
To ensure a thorough analysis, we conduct extensive evaluations on both the MureCom dataset and our custom DreamAssembly dataset.
This dual-dataset testing setup is designed to validate the algorithm's broad adaptability in general scenarios through MureCom, as well as to demonstrate its distinct superiority in specialized contexts through DreamAssembly.
Additionally, to practically assess the visual quality and utility of the data generated by the different methods, we introduce a downstream task evaluation mechanism.
Specifically, for the MureCom dataset, we carefully select two distinct backgrounds for each of the 32 foreground categories to perform the image generation process, ultimately yielding 64 sets of generated images for rigorous metric validation.
Similarly, for the DreamAssembly dataset, we randomly select four different backgrounds for each of the 25 foreground categories to generate a total of 100 sets of images for metric evaluation.
Furthermore, to verify the practical value of the generated images, we conduct an additional experiment exclusively on the DreamAssembly dataset, where we train a YOLOv5 classification model entirely using the data generated by the respective methods. 
Subsequently, we conduct downstream task testing by deploying this trained YOLOv5 model on a separate test set composed entirely of real data.
\begin{figure*}
    \centering
    \includegraphics[width=1\linewidth]{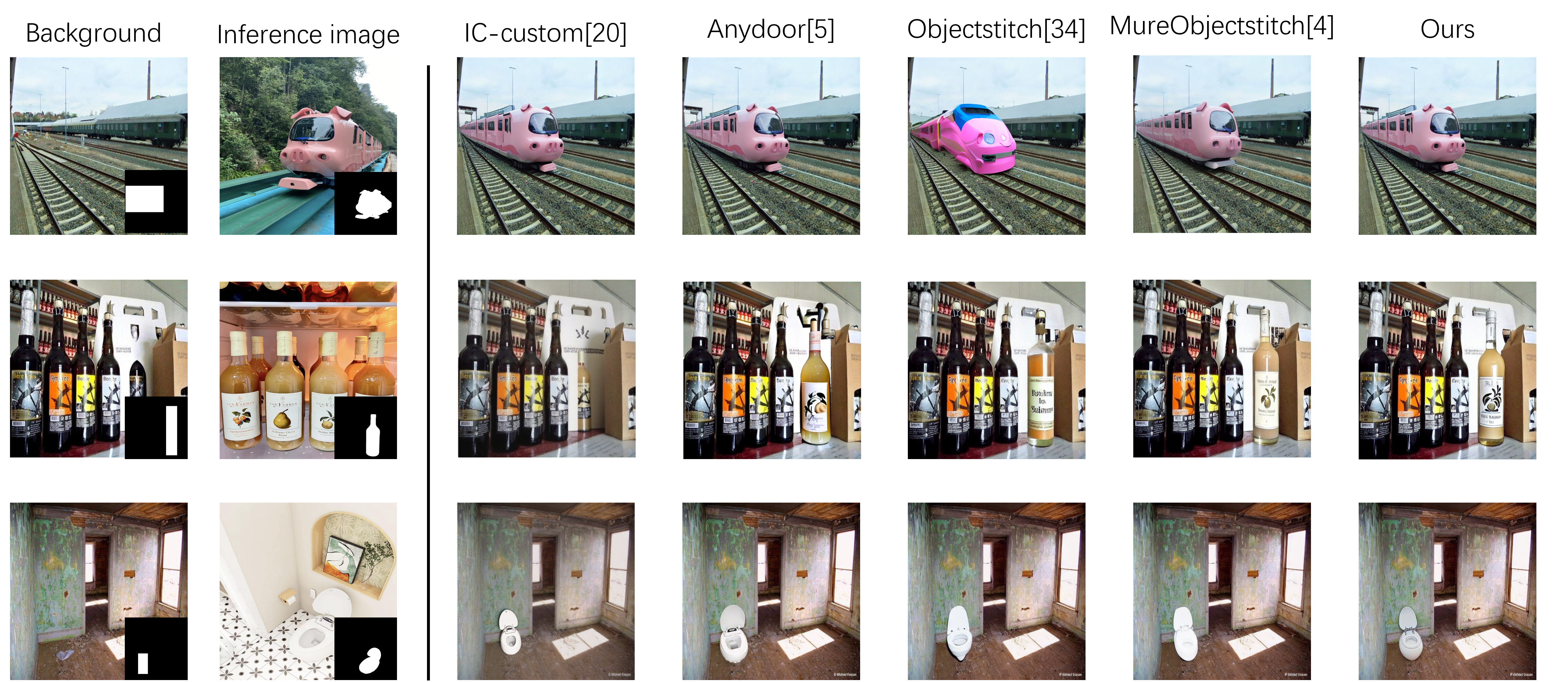}
    \caption{Visualization of different methods on the MureCom dataset. For each row, we present the background image with bounding box and a reference image, followed by the generation outputs of IC-custom, anydoor, Objectstitch, MureObjectstitch and our PostureObjectstitch method.}
    \label{vismurecom}
\end{figure*}
\subsubsection{Quantitative results}

\begin{table}[htbp]
  \caption{Quantitative comparison on MureCom dataset.}
  \label{tab:dreamBooth_comparison}
  \centering
  \resizebox{\linewidth}{!}{
  \begin{tabular}{@{}lcccccc@{}}
    \toprule
    Method & CLIP-I $\uparrow$ & DINO $\uparrow$ &LPIPS $\downarrow$  & SSIM $\uparrow$  \\
    \midrule
    ObjectStitch  & 0.7891 & 0.4831 & 0.0683 & 0.7636  \\
    MureObjectStitch & 0.8139 & 0.5729 & 0.0694 & 0.7633    \\
    Anydoor & 0.8118 & \bf{0.6281} & 0.0676 & \bf{0.8766}   \\
    IC-Custom & 0.7891& 0.5092 & 0.0900 & 0.8452    \\
    Ours & \bf{0.8143} & 0.5885 & \bf{0.0645} & 0.7703  \\
    \bottomrule
  \end{tabular}
  }
\end{table}

\begin{table}[htbp]
  \caption{Quantitative comparison on our Dreamassembly dataset.}
  \label{tab:dreamassembly_comparison}
  \centering
  \resizebox{\linewidth}{!}{
  \begin{tabular}{@{}lcccccc@{}}
    \toprule
    Method & CLIP-I $\uparrow$ & DINO $\uparrow$ &LPIPS $\downarrow$  &SSIM $\uparrow $ \\
    \midrule
    ObjectStitch  & 0.7621 & 0.4103 & 0.0512 & 0.8268  \\
    MureObjectStitch & 0.8033 & 0.5621 & 0.0411 & 0.8366   \\
    Anydoor & \bf{0.8789} & \bf{0.7773} & 0.1677 & 0.8364 \\
    IC-Custom & 0.7011& 0.2258 & 0.4043 & 0.5458    \\
    Ours & 0.8563 & 0.6651 & \bf{0.0230} & \bf{0.9821}  \\
    \bottomrule
  \end{tabular}
  }
\end{table}

\begin{figure*}
    \centering
    \includegraphics[width=1\linewidth]{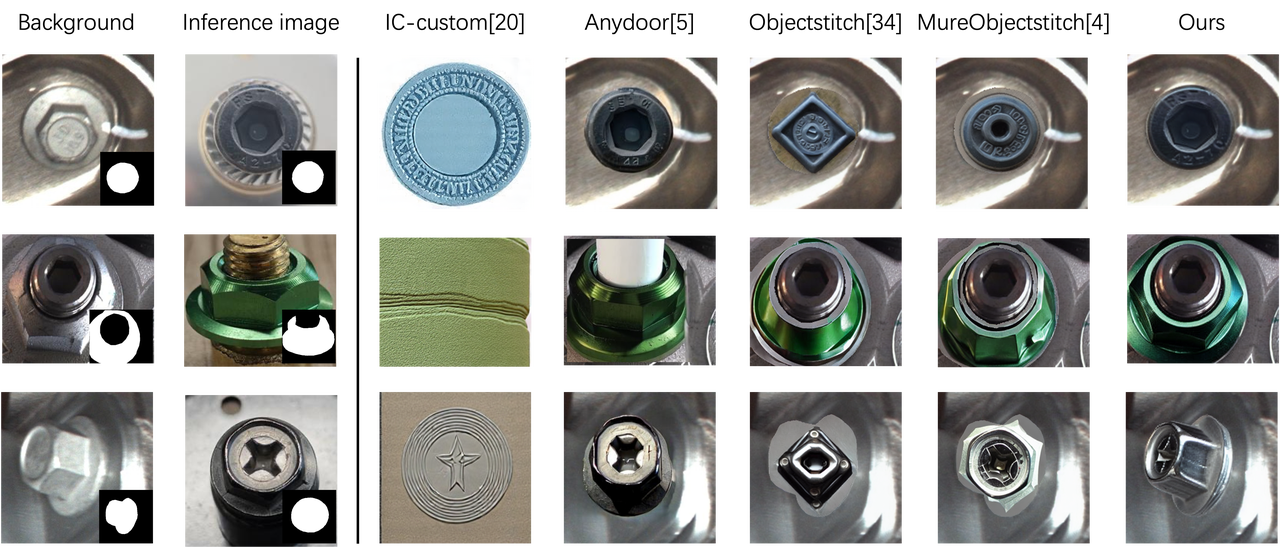}
    \caption{Visualization of different methods on our DreamAssembly dataset. For each row, we present the background image with mask and a reference image, followed by the generation outputs of IC-custom, anydoor, Objectstitch,MureObjectstitch and our PostureObjectstitch method.}
    \label{visassembly}
\end{figure*}

\noindent\cref{tab:dreamBooth_comparison} presents our experimental results on MureCom. Compared to other methods, our approach achieves optimal performance on CLIP-I and LPIPS metrics, while also obtaining competitive results on DINO score and SSIM, demonstrating the adaptability of our method in general domains.
\cref{tab:dreamassembly_comparison} shows our experimental results on DreamAssembly, where we achieve the best performance on both LPIPS and SSIM metrics, indicating that our method excels in background preservation compared to other approaches. For CLIP-I and DINO scores, our method ranks second only to AnyDoor. However, this does not reflect the superiority of our method. In fact, AnyDoor generates foreground images with almost no pose variation relative to the input foreground reference, achieving high CLIP-I and DINO scores at the expense of foreground pose accuracy. Moreover, AnyDoor performs poorly in background preservation, which is also reflected in the LPIPS metric differences. This distinction will be more intuitively demonstrated in the qualitative analysis in the next section.
\cref{tab:yolo} presents the performance of generated data on downstream tasks.
As can be observed, the classification model trained with our generated images achieves superior performance across all metrics compared to other methods, demonstrating that our generated data distribution better approximates the real-world distribution.
\begin{table}[htbp]
  \caption{Quantitative comparison on YOLOv5 classification.}
  \label{tab:yolo}
  \centering
  \resizebox{\linewidth}{!}{
  \begin{tabular}{@{}lcccccc@{}}
    \toprule
    Method & Acc  & Precision & Recall  &F1-score \\
    \midrule
    ObjectStitch  & 0.6667 & 0.7345 & 0.6667 & 0.6991  \\
    MureObjectStitch & 0.6800 & 0.7954 & 0.6800 & 0.7331   \\
    Anydoor & 0.7200 & 0.7909 & 0.7200 & 0.7537 \\
    IC-Custom & 0.2133& 0.2055 & 0.2133 & 0.2093    \\
    Ours & \bf{0.8000} & \bf{0.8437} & \bf{0.8000} & \bf{0.8211}  \\
    \bottomrule
  \end{tabular}
  }
\end{table}

\subsubsection{Qualitative results}
\cref{vismurecom} demonstrates the performance comparison of various methods on the MureCom dataset. These results show that our method exhibits good adaptability to general scenarios.
Additionally, for foreground objects containing textual information, our method achieves even better text restoration quality.
\cref{visassembly} presents the visualization results of different methods on DreamAssembly. 
It can be observed that both AnyDoor and our method perform well in restoring target foreground details, but AnyDoor's text generation capability is inferior to our method. 
ObjectStitch and MureObjectStitch show relatively weaker performance, while IC-Custom's visualization results indicate complete incompatibility with this scenario.
Regarding foreground pose generation and background preservation, AnyDoor fails to adjust foreground object poses according to target backgrounds. Furthermore, for complex assembly relationships such as the nut generation shown in the figure, AnyDoor completely loses background information, producing results that are entirely unsatisfactory. 
Overall, for anomaly generation in industrial scenarios considering assembly relationships, our method achieves the best performance.
\begin{table}[h]
\caption{Ablation Study Results. TFD is target feature distanglement and TM is time-feature modulation. POP means pose and orientation prior. OAL means the OCR auxiliary loss. }
\centering
\footnotesize
\resizebox{\linewidth}{!}{
\begin{tabular}{cccccccc}
\toprule
\multicolumn{4}{c}{\textbf{Method Components}} & \multicolumn{4}{c}{\textbf{Evaluation Metrics}} \\
\cmidrule(lr){1-4} \cmidrule(lr){5-8}
TFD & TM & POP & OAL & CLIP-I $\uparrow$ & DINO $\uparrow$& LPIPS $\downarrow$ & SSIM $\uparrow$ \\
\midrule
 &  &  &  & 0.8033 & 0.5621 & 0.0411 & 0.8366 \\
\checkmark &  &  &  & 0.8212 & 0.6012 & 0.0422 & 0.8582 \\
\checkmark & \checkmark &  &  & 0.8401 & \bf{0.6715} & 0.0485 & 0.8802 \\
\checkmark & \checkmark & \checkmark &  & 0.8425 & 0.6612 & \bf{0.0225} & 0.9815 \\
\checkmark & \checkmark & \checkmark & \checkmark & \bf{0.8563} & 0.6651 & 0.0230 & \bf{0.9821}\\
\bottomrule
\end{tabular}
}
\label{tab:ablation}
\end{table}
\subsection{Ablation Studies}
To rigorously validate the individual contributions and effectiveness of our proposed components, we design a series of comprehensive ablation studies for verification.
All ablation experiments are exclusively conducted on the DreamAssembly dataset, with the detailed quantitative comparative results systematically presented in the corresponding table.
By comparing the results in the first and second rows, we clearly observe noticeable improvements in both the CLIP-I and DINO evaluation scores following the implementation of feature decoupling.
This indicates that the feature decoupling mechanism successfully isolates relevant attributes and has a distinctly positive impact on the overall generation quality.
Furthermore, when comparing the second and third rows, we find that applying time modulation to these decoupled features leads to a significant improvement in the DINO scores, which effectively demonstrates the critical role and effectiveness of this specific component.
Subsequently, after incorporating the positional pose prior information, the DINO scores experience a slight decrease, whereas the LPIPS and SSIM metrics show significant improvements.
This phenomenon occurs because the introduced prior information enables the model to actively adjust the pose of the foreground objects.
Consequently, this adjustment causes some structural variations compared to the original input foreground, which inevitably results in slightly lower DINO scores.
Finally, after adding the OCR auxiliary loss to the training objective, no particularly significant increases are observed across the quantitative metrics.
However, despite the modest quantitative gains, the positive effect of this OCR loss is highly significant in the qualitative visualization results.
More detailed visual comparisons regarding this component will be presented in the appendix.

\section{Conclusion}
In summary, this study introduces PostureObjectStitch, a novel anomaly image generation method specifically tailored for the complex demands of industrial assembly scenarios.
Specifically, by systematically decoupling foreground features into distinct RGB, high-frequency, and texture components, our approach can precisely capture fine-grained object characteristics.
Simultaneously, it adaptively optimizes the utilization of these decoupled features across different timesteps of the diffusion model.
Furthermore, the integration of conditional loss and position-pose prior mechanisms plays a crucial role in ensuring strict semantic consistency and spatial rationality in the final generation results.
To comprehensively validate the effectiveness of our proposed method, we construct and introduce the DreamAssembly dataset, which is specifically designed as a dedicated benchmark for evaluating industrial assembly anomaly generation.
Extensive experimental results demonstrate that PostureObjectStitch significantly outperforms existing general methods in terms of both complex assembly relationship modeling and high-resolution detail fidelity, thereby offering a highly robust and practical solution for downstream industrial quality control applications.
Looking ahead to future work, we plan to integrate multi-modal data sources, including 3D models and point cloud data, to conduct a deeper and more comprehensive investigation into assembly relationship generation.
Through these advancements, we expect to generate more realistic and strictly physically consistent assembly anomalies that can better reflect the complex dynamics of real-world industrial scenarios.
Ultimately, we believe that this line of research will pave the way for more intelligent, efficient, and automated visual inspection systems in the era of smart manufacturing.

\appendix
\clearpage
\setcounter{page}{1}

\bibliographystyle{ACM-Reference-Format}
\bibliography{sample-base}










\end{document}